\documentclass[12pt,a4paper]{elex}
\usepackage[left=2.5cm,right=2.5cm,bottom=2.5cm,top=2.5cm]{geometry}
\usepackage[english]{babel}
\usepackage[T1]{fontenc}
\usepackage[utf8]{inputenc}
\usepackage[unicode=true]{hyperref}
\usepackage[font=small]{caption}
\usepackage{graphicx}
\usepackage{natbib}

\usepackage{amssymb}        % added by XM for alpha beta greek letters
\usepackage{glossaries}     % added by XM for acronym.tex
\usepackage{tikz}           % aded by XM
\usetikzlibrary{shapes.geometric, arrows} % added by XM for LATEX figures
\usepackage{pgfplotstable}  % added by XM for LATEX plots
\usepackage{tipa}           % added by XM for IPA letter
\newacronym{ann}{ANN}{Artificial Neural Network}
\newacronym{asr}{ASR}{Automatic Speech Recognition}
\newacronym{conv1d}{Conv1D}{1D convolution layer}
\newacronym{ctc}{CTC}{Connectionist Temporal Classification}
\newacronym{gipfa}{GIPFA}{Generating IPA Pronunciation From Audio}
\newacronym{ipa}{IPA}{International Phonetic Alphabet}
\newacronym{ll}{LL}{LinguaLibre}
\newacronym{lstm}{LSTM}{Long Short Term Memory}
\newacronym{mfcc}{MFCC}{Mel-Frequency Cepstral Coefficients}
\newacronym{nlp}{NLP}{Natural Language Processing}
\newacronym{nmt}{NMT}{Neural Machine Translation}
\newacronym{std}{std}{standard deviation}
\newacronym{url}{URL}{Uniform Resource Locator}
\newacronym{wav}{WAV}{Waveform Audio File}
\newacronym{wer}{WER}{Word Error Rate}       % added by XM
\usepackage{siunitx}        % added by XM for numbers like 1e-4

\bibpunct[: ]{(}{)}{;}{a}{,}{,}

\let\subparagraph\paragraph

\usepackage{enumitem}
\setlist{leftmargin=3em}

\makeatletter
\renewcommand{\@seccntformat}[1]{%
  \ifcsname prefix@#1\endcsname
    \csname prefix@#1\endcsname
  \else
    \csname the#1\endcsname\quad
  \fi}
\newcommand\prefix@section{\thesection. }
\makeatother

\begin{document}
\mainmatter
\title{GIPFA: Generating IPA Pronunciation from Audio}
\titlerunning{GIPFA}
\author{\bf Xavier Marjou}
\institute{Lannion, Brittany, France\\
E-mail: xavier.marjou@gmail.com}
\toctitle{Title for TOC}

\maketitle

\begin{abstract}
Transcribing spoken audio samples into \gls{ipa} has long been reserved for experts. In this study, we examine the use of an \gls{ann} model to automatically extract the \gls{ipa} phonemic pronunciation of a word based on its audio pronunciation, hence its name \gls{gipfa}. Based on the French Wikimedia dictionary, we trained our model which then correctly predicted $75\%$ of the IPA pronunciations tested. Interestingly, by studying inference errors, the model made it possible to highlight possible errors in the dataset as well as identifying the closest phonemes in French.

\end{abstract}

\section{Introduction}

\glsresetall %% Added by XMs

Some dictionaries like Wiktionary offer both listening to words spoken by real users and reading  phonemic pronunciations in the form of the \gls{ipa}. 

However, in the case of the French Wiktionary, the phonemic \gls{ipa} transcripts are subject to a small percentage of errors. Several reasons can explain these errors. First, Wiktionary contributors may not be \gls{ipa} experts; second, even \gls{ipa} experts sometimes may make careless mistakes; third, the audio may be inconsistent because it is generally recorded independently without taking \gls{ipa} pronunciation into account, which can lead to important discrepancies; fourth, some sounds like \textipa{/o/} and \textipa{/O/} may be very close to each other and can depend on the speaker.

This article examines whether such errors could be avoided by using an \gls{nlp} tool to automatically extract phonemic \gls{ipa} pronunciation from audio pronunciation.

To this purpose, we made use of \gls{asr}, which has already been the subject of in-depth studies. In particular, many recent implementation approaches have successfully used a deep \gls{ann} as in \cite{han2020contextnet} and \cite{das2019advancing}, hence our choice to design a new \gls{ann} called \gls{gipfa}. In order to train and test it, we also assembled a new experimental dataset based on $80400$ samples from the French Wiktionary.

Despite a dataset containing an unknown percentage of erroneous data samples, our \gls{gipfa} model succeeded in providing reasonable accuracy. Although it failed to replace \gls{ipa} experts, it nevertheless proved to be particularly useful in identifying the biggest errors in the dataset.

\section{Methodology} \label{methodology} 

In order to predict the \gls{ipa} pronunciation of a word, two main steps were necessary: identifying a relevant dataset and designing an \gls{ann} model capable of inferring an \gls{ipa} pronunciation from an audio pronunciation.

\subsection{Dataset}

\begin{table*}[h!]
  \begin{center}
    \begin{tabular}{c c c} % <-- Alignments: 1st column left, 2nd middle and 3rd right, with vertical lines in between
      \hline
      Word & Audio filename & IPA pronunciation \\
      \hline
      bonjour & LL-Q150\ (fra)-LoquaxFR-bonjour.wav & \textipa{b\~OZuK} \\
      \hline
      \newline
    \end{tabular}
    \caption{Dataset}
    \label{tab:dataset_features}
  \end{center}
\end{table*}

Our dataset came from a Wikimedia dump\footnote{\url{https://dumps.wikimedia.org/frwiktionary/20200501/}} containing all pages and articles of the French Wiktionary. In this dump, each page generally contains three essential features: one \emph{word} along with $n$ main \emph{\gls{ipa} pronunciations} and $m$ examples of \emph{audio pronunciations} recorded by several speakers. 

\begin{itemize}
\item A word is a text string containing Unicode characters. The \emph{word} terminology has to be taken in the broad sense as a Wiktionary word contains common names, proper names words, abbreviations, numbers and even sayings. Although our \gls{ann} did not use it, we kept the word in our dataset for debugging purposes, in order to have the possibility to find back the Wiktionary page containing the pronunciations. 

\item An audio pronunciation refers to an audio file generally recorded in a \gls{wav} format containing the pronounced word. Wiktionary pages can contain one or more audio pronunciations for the same word. When an audio file is generated with \gls{ll}\footnote{\url{https://lingualibre.org}} software, it benefits from three useful features: the audio file is under the Creative Commons sharing license\footnote{\url{https://creativecommons.org/licenses/by-sa/4.0/}}; the file can be fetched from Wikimedia Commons\footnote{\url{https://commons.wikimedia.org/}} based on its audio filename; the audio filename also contains a label representing a user name which can be used to identify audio files generated by users. 

\item An \gls{ipa} pronunciation is a text string containing \gls{ipa} symbols. For learning purposes, each audio pronunciation of a word should ideally be associated to a single \gls{ipa} pronunciation transcribing this precise audio content; a ranking of the most common pronunciations might also be calculated and indicated in the page describing the word. However, most words have a single \gls{ipa} pronunciation (i.e. $n=1$) even when multiple audio pronunciations are available. Although some words have multiple \gls{ipa} pronunciations (e.g. \emph{coût}), a Wiktionary page rarely indicates which of these pronunciations corresponds to an audio file. 

\end{itemize}

For our purpose, we restricted our dataset to samples containing:
\begin{itemize}
    \item Words of the French Wiktionary\footnote{\url{https://fr.wiktionary.org/}};
    \item French words, given that each Wiktionary describes words of several languages;
    \item Words with a single \gls{ipa} pronunciation, given that multiple \gls{ipa} per audio sample introduce ambiguities;
    \item \gls{ipa} pronunciation containing symbols making part of the $37$ traditional French phonemes (i.e. '\textipa{i}', '\textipa{e}', '\textipa{E}', '\textipa{a}', '\textipa{A}', '\textipa{O}', '\textipa{o}', '\textipa{u}', '\textipa{y}', '\textipa{\o}', '\textipa{\oe}', '\textipa{@}', '\textipa{\~E}', '\textipa{\~A}', '\textipa{\~O}', '\textipa{\~\oe}', '\textipa{j}', '\textipa{w}', '\textipa{4}', '\textipa{p}', '\textipa{k}', '\textipa{t}', '\textipa{b}', '\textipa{d}', '\textipa{g}', '\textipa{f}', '\textipa{s}', '\textipa{S}', '\textipa{v}', '\textipa{z}', '\textipa{Z}', '\textipa{l}', '\textipa{K}', '\textipa{m}', '\textipa{n}', '\textipa{\textltailn}', '\textipa{N}');
   \item \gls{ipa} pronunciation containing less than $20$ phonemes, in order to keep our \gls{ann} model reasonable in size regarding our resources;
    \item Audio files recorded with \gls{ll}, in order to easily fetch audio files. 
\end{itemize}

We also discarded 9 symbols that appear as optional in the \gls{ipa} pronunciation of the French Wiktionary ('\textipa{\t{}}', '.', ' ', '\textipa{\t*{}}', '\textipa{'}' and '\textipa{:}', '(', ')', '-').

The resulting dataset contained $80200$ samples from $102$ different speakers. As depicted in Table \ref{tab:dataset_features}, each sample contained three features: a \emph{word}, an \emph{audio filename} and an \emph{IPA pronunciation}. 

In addition, we also pre-processed the \gls{wav} files to have a fix length of $2$ seconds, and then converted them into an \gls{mfcc} format so that they could serve as direct inputs into our model. Although processing audio files under a \gls{wav} format would be possible as in \cite{43960}, it requires significant RAM memory, hence our choice to transpose them into a \gls{mfcc} format, as usually performed in many studies like in \cite{alcaraz2009speech} and  \cite{nahid2017bengali}.

\subsection{Experiments}

\subsubsection{Model architecture}

\begin{figure*}
    \centering
\begin{tikzpicture}
\definecolor{myolive}{rgb}{0.8,0.725,0.454}
\definecolor{myviolet}{rgb}{0.505,0.447, 0.701}
\tikzstyle{data_} = [rectangle, minimum width=0.1cm, minimum height=0.1cm,text centered, draw=black, fill=myolive!30]
\tikzstyle{layer_} = [rectangle, minimum width=0.1cm, minimum height=0.1cm,text centered, draw=black, fill=myviolet!30]
\tikzstyle{arrow} = [thick,->,>=stealth]
\node (data1) [data_, align=center] {Audio \\ data};
\node (layer1) [layer_,right of=data1, xshift=0.75cm, rotate=90] {Conv1d};
\node (layer2) [layer_,right of=layer1, rotate=90] {Relu};
\node (layer3) [layer_,right of=layer2, rotate=90] {Conv1d};
\node (layer4) [layer_,right of=layer3, rotate=90] {Relu};
\node (layer5) [layer_,right of=layer4, rotate=90] {LSTM};
\node (layer6) [layer_,right of=layer5, rotate=90] {LSTM};
\node (layer7) [layer_,right of=layer6, rotate=90] {Linear};
\node (data2) [data_, right of=layer7, xshift=0.75cm, align=center] {IPA \\ data};
\draw [arrow] (data1) -- (layer1);
\draw [arrow] (layer1) -- (layer2);
\draw [arrow] (layer2) -- (layer3);
\draw [arrow] (layer3) -- (layer4);
\draw [arrow] (layer4) -- (layer5);
\draw [arrow] (layer5) -- (layer6);
\draw [arrow] (layer6) -- (layer7);
\draw [arrow] (layer7) -- (data2);
\node (module1) [label] {};
\end{tikzpicture}
\caption{The GIPFA ANN model used for transcribing audio samples into IPA samples.}
\label{fig:gipfa_ann}
\end{figure*}
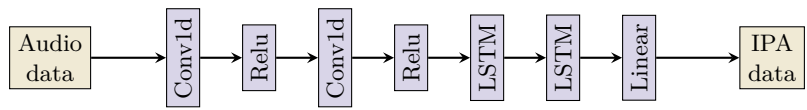

We modeled our \gls{gipfa} \gls{ann} as depicted in Figure \ref{fig:gipfa_ann}. It contains typical components found in many \gls{ann} models used for \gls{asr}. However, given that we only had to translate a single word per sample, we did not use any Transformer component \citep{vaswani2017attention}. Each audio input sample (\gls{mfcc} data) first traversed a stack of two \gls{conv1d} layers to extract the shape of the \gls{mfcc} data; followed by two \gls{lstm} filters \citep{hochreiter1997long} to extract temporal sequences; and finally followed by a linear layer in order to allow a \gls{ctc} loss calculation \citep{graves2012connectionist}. We did not allow the succession of two identical phonemes because this is rare in French words. In addition, we used an AdamW optimizer \citep{loshchilov2017decoupled} with a learning rate of \num{1e-4}.

\subsubsection{Hyperparameters}

We used Ray Tune \citep{moritz2018ray} for fine-tuning our hyperparameters with respect to accuracy results. It led us to identify a set of best values among a larger set of experimented values as summarized in Table \ref{tab:hyperparameters}. The resulting model contained \num{9609558} trainable parameters. Slight variations in the best values did not lead to significant improvement. Although it is believed that wider network may have lead to better results \citep{nakkiran2019deep}, we limited our model to these $10M$ parameters due to our limited computing resources. 

\begin{table}[h!]
  \begin{center}
    \begin{tabular}{c c c } % <-- Alignments: 1st column left, 2nd middle and 3rd right, with vertical lines in between
      \hline
      Hyperparameter & Tested values & Best value  \\
      \hline
      mfcc\_coefficients & 40 & $40$ \\
      conv1d\_activ & none, relu & relu  \\
      conv1d\_layers & $0$, $1$, $2$, $3$ & $2$  \\
      conv1d\_units & $32$, $64$, $128$ & $128$  \\
      conv1d\_bn & False, True & True  \\
      lstm\_layers & $0$, $1$, $2$ & $2$  \\
      lstm\_units & $128$, $256$, $512$ & $512$  \\
      lstm\_dropout & $0.1$, $0.25$, $0.5$ & $0.5$  \\
      lstm\_bidir & False, True & True  \\ 
      lstm\_bn & False, True & True  \\  
      optimizer & Adam, AdamW & AdamW  \\   
      lr & 1e-3, 1e-4 & 1e-4  \\   
      \hline
      \newline
    \end{tabular}
    \caption{GIPFA hyperparameters values}
    \label{tab:hyperparameters}
  \end{center}
\end{table}

\subsubsection{Training}

For the training step, we used \num{79326} samples distributer over \num{3966} batches of \num{20} samples (\num{3927} training batches and \num{39} evaluation batches). During a pre-processing step, all audio samples were standardized with a the mean ($-11.48$) and standard deviation ($80.30$) pre-observed on the dataset. 

Before each run, the data samples were randomly shuffled. Each training run took approximately 10 epochs of 3 minutes each on a single GPU (GeForce RTX 2080, 8 GB). 

\subsubsection{Test}

For the testing step, we used \num{1000} unseen samples to evaluate the performances of the \gls{gipfa} \gls{ann}. 

\subsubsection{Accuracy}

Since solving the translation problem requires correct inference of the entire \gls{ipa} pronunciation, we simply set for each tested sample an accuracy of $1$ when our model predicted an \gls{ipa} pronunciation equal to the tested target \gls{ipa} pronunciation, or $0$ otherwise. After each training run, we then calculated the average accuracy across all samples (i.e. average accuracy between \num{0.0} and \num{1.0}).

We performed 11 runs (with one training step and one test step for each) to allow reasonable confidence in the average accuracy results. We finally computed a mean accuracy and the associated \gls{std} on the 11 tests. 

Since the dataset had not been studied further, there was unfortunately no baseline reference to challenge our results.

\subsubsection{Enlightenment on errors}

To our knowledge, no study has examined the exactness and coherence of the audio files and \gls{ipa} pronunciations of the French Wiktionary, meaning that the dataset may contain errors, making it difficult to assess whether a prediction error comes from the dataset or from the \gls{ann}.

In order to obtain more in-depth information on errors, we therefore also calculated three other metrics related to the $80000$ samples in the dataset:
\begin{itemize}
    \item At the word level
    \begin{itemize}
    \item \emph{Edit distance error}: the Levenshtein distance \citep{levenshtein1965} between the predicted \gls{ipa} pronunciation and the target \gls{ipa} pronunciation, in order to estimate how far the prediction was from the target.
    \end{itemize}
    \item At the phoneme level
    \begin{itemize}
    \item \emph{Average phoneme accuracy}: the percentage of correct translations for each phoneme;
     \item \emph{Error pair percentage}: Since each of the 37 target phonemes can be incorrectly translated as one of the other 36 phonemes, the results can contain up to 37 * 36 categories of error pairs. To assess the representativeness of each pair, we calculated its number of occurrences divided by the total number of phonemic errors.
     \end{itemize}
\end{itemize}

The code is available on Github \footnote{Code available at \url{https://github.com/marxav/gipfa}}.

\section{Results} \label{results}

In this section, we describe two different results: first, the accuracy of the model; then a more detailed observation of errors at the phoneme level and at the word level.

\subsection{Accuracy}

\begin{table}[h!]
  \begin{center}
    \begin{tabular}{c c c c} % <-- Alignments: 1st column left, 2nd middle and 3rd right, with vertical lines in between
      \hline
      Training samples & Tested samples & Pronunciation accuracy & Pronunciation accuracy   \\
       &  & (mean) & (std)  \\
      \hline
      $79326$ & $1000$ & $0.75$ & $0.02$ \\
      \hline
      \newline
    \end{tabular}
    \caption{Pronunciation accuracy}
    \label{tab:accuracy_results}
  \end{center}
\end{table}

Table \ref{tab:accuracy_results} presents the accuracy results which were consistent consistent across the 11 runs; our \gls{gipfa} \gls{ann} model successfully predicted around \num{75} \gls{ipa} pronunciations out \num{100} audio samples.

Correctly inferred pronunciations had a mean length of 7.51 whereas incorrectly inferred pronunciations had a mean length of 8.65 thus indicating a slightly higher probability of error as the length of the \gls{ipa} pronunciation increased.

\subsection{Insights on the errors}

Performing inferences on $80000$ samples of the dataset allowed to better understand the reasons for the errors.

\subsubsection{Phoneme Accuracy}

Table \ref{tab:accuracy_per_phoneme} reports the translation accuracy of each phoneme. One phoneme (/\textipa{A}/) had poor accuracy (less than $50\%$), five phonemes (/\textipa{o}/, /\textipa{N}/, /\textipa{\~\oe}/,  /\textipa{\textltailn}/ and /\textipa{oe}/) had moderate accuracy  (between $65\%$  and $89\%$) while the remaining thirty-one phonemes had high accuracy (over $90\%$). 

     \begin{table}[h!]
    \begin{center}
    \begin{tabular}{c c c c} 
      \hline
      Target & Correct & Incorrect & Average \\
      phoneme & translation & translation &  accuracy\\
      \hline
          \textipa{A} & \num{392} & \num{605} & \num{0.39} \\
      \textipa{o} & \num{4,615} & \num{2485} & \num{0.65} \\
      \textipa{N} & \num{40} & \num{17} & \num{0.70} \\
      \textipa{\~\oe} & \num{241} & \num{89} & \num{0.73} \\
      \textipa{\textltailn} & \num{697} & \num{110} & \num{0.86} \\
      \textipa{\oe} & \num{2459} & \num{301} & \num{0.89} \\
      \textipa{4} & \num{1185} & \num{113} & \num{0.91} \\
      \textipa{E} & \num{15859} & \num{1472} & \num{0.92} \\
      \textipa{@} & \num{7918} & \num{732} & \num{0.92} \\
      \textipa{g} & \num{5911} & \num{427} & \num{0.93} \\
      \textipa{\o} & \num{2587} & \num{169} & \num{0.94} \\
      \textipa{O} & \num{18655} & \num{1074} & \num{0.95} \\
      \textipa{e} & \num{30018} & \num{1608} & \num{0.95} \\
      \textipa{w} & \num{4357} & \num{159} & \num{0.96} \\
      \textipa{v} & \num{7469} & \num{282} & \num{0.96} \\
      \textipa{u} & \num{6712} & \num{250} & \num{0.96} \\
      \textipa{\~E} & \num{4527} & \num{192} & \num{0.96} \\
      \textipa{j} & \num{12567} & \num{547} & \num{0.96} \\
      \textipa{b} & \num{12753} & \num{434} & \num{0.97} \\
      \textipa{n} & \num{13165} & \num{472} & \num{0.97} \\
      \textipa{p} & \num{14845} & \num{464} & \num{0.97} \\
      \textipa{l} & \num{23181} & \num{684} & \num{0.97} \\
      \textipa{\~A} & \num{13704} & \num{226} & \num{0.98} \\
      \textipa{f} & \num{9632} & \num{225} & \num{0.98} \\
      \textipa{y} & \num{8235} & \num{183} & \num{0.98} \\
      \textipa{z} & \num{7730} & \num{146} & \num{0.98} \\
      \textipa{i} & \num{34772} & \num{664} & \num{0.98} \\
      \textipa{d} & \num{15975} & \num{323} & \num{0.98} \\
      \textipa{k} & \num{23159} & \num{503} & \num{0.98} \\
      \textipa{S} & \num{4407} & \num{92} & \num{0.98} \\
      \textipa{a} & \num{44575} & \num{707} & \num{0.98} \\
      \textipa{m} & \num{17334} & \num{313} & \num{0.98} \\
      \textipa{K} & \num{47221} & \num{799} & \num{0.98} \\
      \textipa{Z} & \num{5552} & \num{137} & \num{0.98} \\
      \textipa{t} & \num{29691} & \num{713} & \num{0.98} \\
      \textipa{\~O} & \num{9258} & \num{129} & \num{0.99} \\
      \textipa{s} & \num{30018} & \num{400} & \num{0.99} \\
      \hline
      \newline
    \end{tabular}
    \caption{Average accuracy of each phoneme}
    \label{tab:accuracy_per_phoneme}
    \end{center}
    \end{table}
    
\begin{figure*}
\begin{center}
\includegraphics[scale=0.5]{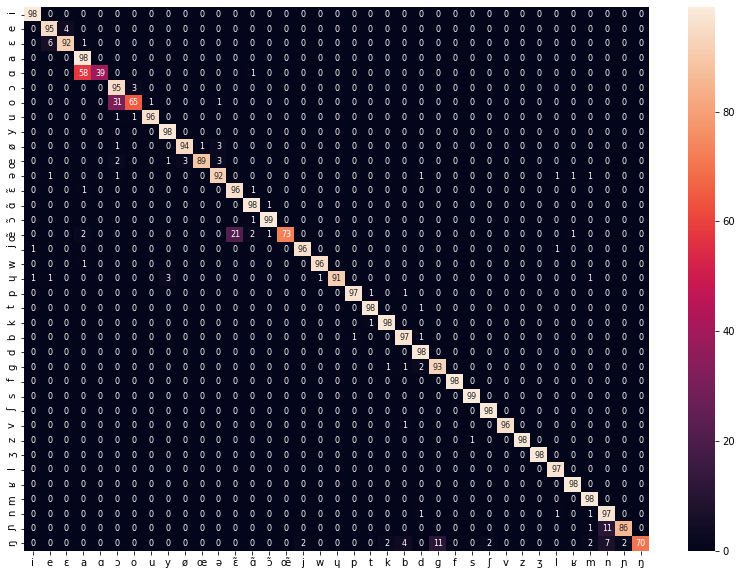}
\caption{Confusion Matrix}
\label{fig:confusion-matrix}
\end{center}
\end{figure*}

To better observe the details, we also detailed these phoneme translation errors in a confusion matrix as shown in Figure \ref{fig:confusion-matrix}. Each row in the matrix represented a target phoneme while each column represented the distribution of predicted phonemes. For instance, it turned out that the target phoneme /\textipa{E}/ was $6\%$ of the time predicted as /\textipa{e}/, $92\%$ as /\textipa{E}/ and $1\%$ /\textipa{a}/. Notable outliers were four large numbers outside the diagonal: $58\%$ of /\textipa{A}/ seemed poorly predicted as an /\textipa{a}/; $31\%$ of /\textipa{o}/ as /\textipa{O}/; $21\%$ of /\textipa{\~\oe}/ as /\textipa{\~E}/; and $11\%$ of /\textipa{N}/ as /\textipa{g}/; It turned out that, like humans, the \gls{ann} had difficulties in differentiating near elementary sounds.

\subsubsection{Error pair percentage}

Table \ref{tab:topmost_bad_predictions} represents the proportion of the error associated with each phoneme pair compared to the total errors of all pairs of phonemes. Interestingly, only three pairs of phonemes generated $31\%$ of all errors: (/\textipa{o}/, /\textipa{O}/) ($15\%$ of all errors), (/\textipa{e}/, /\textipa{E}/) ($12\%$ of all errors) and (/\textipa{a}/, /\textipa{A}/) ($4\%$ of all errors).

\begin{table}[h!]
    \begin{center}
    \begin{tabular}{c c c} 
      \hline
      Target & Predicted & Percentage of  \\
      phoneme & phoneme & all errors  \\
      \hline
      \textipa{o} & \textipa{O} & $12.03\%$ \\
      \textipa{e} & \textipa{E} & $6.51\%$ \\
      \textipa{E} & \textipa{e} & $5.46\%$ \\
      \textipa{A} & \textipa{a} & $3.16\%$ \\
      \textipa{O} & \textipa{o} & $3.07\%$ \\
      \textipa{t} & \textipa{d} & $1.25\%$ \\
      \textipa{E} & \textipa{a} & $1.04\%$ \\
      \textipa{a} & \textipa{A} & $0.83\%$ \\
      \hline
      \newline
    \end{tabular}
    \caption{Most encountered error pairs}
    \label{tab:topmost_bad_predictions}
    \end{center}
    \end{table}

\subsubsection{Word-level distance error}

\begin{table}[h!]
  \begin{center}
    \begin{tabular}{c c } % <-- Alignments: 1st column left, 2nd middle and 3rd right, with vertical lines in between
      \hline
      Computed & Levenshtein distance   \\
      samples & mean, std  \\
      \hline
      $80000$ & $0.31$, $0.66$ \\
      \hline
      \newline
    \end{tabular}
    \caption{Levenshtein distance}
    \label{tab:Levenshtein_results}
  \end{center}
\end{table}

Table \ref{tab:Levenshtein_results} reports a small mean Levenshtein distance and gives assurance that there is strong consistency between the audio content and the \gls{ipa} pronunciation for the samples in the dataset studied.

    \begin{table*}[h!]
    \begin{center}
    \begin{tabular}{c c c c} 
      \hline
      Word & \gls{ipa} Target & \gls{ipa} Prediction & Levenshtein distance \\
      \hline
          1337 & /\textipa{l}\textipa{i}\textipa{t}/ & /\textipa{m}\textipa{i}\textipa{t}\textipa{a}\textipa{s}\textipa{\~A}\textipa{t}\textipa{K}\textipa{\~A}\textipa{m}\textipa{z}\textipa{O}\textipa{t}/ & 13 \\
      agent innervant & /\textipa{a}\textipa{Z}\textipa{\~A}\textipa{i}\textipa{n}\textipa{E}\textipa{K}\textipa{v}\textipa{\~A}/ & /\textipa{g}\textipa{o}/ & 11 \\
      brut de décoffrage & /\textipa{b}\textipa{K}\textipa{y}\textipa{t}\textipa{d}\textipa{@}\textipa{d}\textipa{e}\textipa{k}\textipa{O}\textipa{f}\textipa{K}\textipa{a}\textipa{Z}/ & /\textipa{s}\textipa{b}\textipa{O}\textipa{K}\textipa{d}\textipa{e}\textipa{d}\textipa{t}\textipa{O}\textipa{K}/ & 10 \\
      Michel & /\textipa{m}\textipa{i}\textipa{S}\textipa{E}\textipa{l}/ & /\textipa{s}\textipa{t}\textipa{\~E}\textipa{d}\textipa{@}\textipa{s}\textipa{\~A}\textipa{m}\textipa{S}\textipa{E}\textipa{l}/ & 10 \\
      phalange proximale & /\textipa{f}\textipa{a}\textipa{l}\textipa{\~A}\textipa{Z}\textipa{p}\textipa{K}\textipa{O}\textipa{k}\textipa{s}\textipa{i}\textipa{m}\textipa{a}\textipa{l}/ & /\textipa{f}\textipa{a}\textipa{l}\textipa{\~A}\textipa{Z}/ & 9 \\
      analyse calorimétrique & /\textipa{a}\textipa{n}\textipa{a}\textipa{l}\textipa{O}\textipa{g}\textipa{S}\textipa{i}\textipa{m}\textipa{i}\textipa{k}/ & /\textipa{a}\textipa{n}\textipa{a}\textipa{l}\textipa{i}\textipa{s}\textipa{k}\textipa{a}\textipa{l}\textipa{O}\textipa{K}\textipa{i}\textipa{m}\textipa{e}\textipa{t}\textipa{i}\textipa{k}/ & 9 \\
      àtha & /\textipa{a}\textipa{t}\textipa{\~O}\textipa{n}\textipa{\~\oe}\textipa{b}\textipa{l}\textipa{a}\textipa{v}\textipa{i}/ & /\textipa{a}\textipa{t}\textipa{a}/ & 9 \\
      Wikitionnaire & /\textipa{g}\textipa{a}\textipa{z}\textipa{a}\textipa{e}\textipa{f}\textipa{E}\textipa{d}\textipa{@}\textipa{s}\textipa{f}\textipa{E}\textipa{K}/ & /\textipa{g}\textipa{O}\textipa{Z}\textipa{i}\textipa{f}\textipa{i}\textipa{s}\textipa{\o}\textipa{l}\textipa{E}\textipa{K}/ & 9 \\
      arrondir par défaut & /\textipa{a}\textipa{K}\textipa{\~O}\textipa{d}\textipa{i}\textipa{K}\textipa{p}\textipa{a}\textipa{K}\textipa{d}\textipa{e}\textipa{f}\textipa{o}/ & /\textipa{a}\textipa{K}\textipa{\~A}\textipa{d}\textipa{i}\textipa{K}/ & 8 \\
      Luxembourg & /\textipa{l}\textipa{y}\textipa{k}\textipa{s}\textipa{\~A}\textipa{b}\textipa{u}\textipa{K}/ & /\textipa{y}\textipa{s}\textipa{e}\textipa{K}\textipa{z}\textipa{O}\textipa{n}\textipa{b}/ & 8 \\
      \hline
      \newline
    \end{tabular}
    \caption{Top-10 pronunciations with the highest Levenshtein distance}
    \label{tab:top_levenshtein_distances}
    \end{center}
    \end{table*}
    
However, Table \ref{tab:top_levenshtein_distances} focuses on the most extreme outliers by reporting the 10 samples with the highest Levenshtein distance. Upon investigation, it was found that all these $10$ samples contained either an error in the audio sample (e.g. bad word spoken or no word spoken at all) or an error in the target \gls{ipa} pronunciation, which meant that all these errors were in the dataset itself. These results therefore suggest that data samples whose pronunciations have a high Levenshtein distance probably contain an error.

Additional work would be required to identify the best threshold distance to identify possible errors in the dataset.

\section{Discussion and Conclusion}

Previous work has documented the effectiveness of the \gls{ann} model for \gls{asr}. However most studies have focused on the direct translation of audio samples into words. 

In this study, we focused instead on the translation of audio samples into phonemes. We first proposed an \gls{ann} predicting with $75\%$ accuracy the French pronunciations of the French Wiktionary. 

Since to our knowledge no existing work has been done on this specific task and dataset, there was no basis for comparison or assurance as to the accuracy and consistency of the data. 

We have shown that the translations of certain phonemes were more problematic since some phonemes are close elementary sounds (\textipa{/o/} and \textipa{/O/}, \textipa{/E/} and \textipa{/e/}, \textipa{/A/} and \textipa{/a/}) and thus difficult to be distinguished. Future work may consider carefully checking the audio samples and \gls{ipa} pronunciations containing these close phonemes, which would in turn enhance the efficiency of the \gls{ann}. In addition, future work could also involve synthesized audio examples and use them as additional samples to reinforce training data.

However, we have also shown that the Levenshtein distance between our \gls{gipfa} prediction and the target (as it exists in the dataset and therefore in the Wiktionary) can highlight the most suspect samples in the dataset. Such results therefore suggest that our \gls{gipfa} \gls{ann} would be a valuable tool to help verify  the consistency of the Wiktionary regarding pronunciation.

Therefore, integrating it into a tool like \gls{ll} should be useful in order to suggest an \gls{ipa} transcription. It could even be used to suggest an \gls{ipa} transcription associated with each recorded audio sample, since having one \gls{ipa} transcription per audio file should further improve the performances of the \gls{ann}.

Finally, we believe this method should be applicable to other languages provided that a sufficient number of training samples are available.

\section*{Acknowledgements}
We thank all Wiktionary and LinguaLibre contributors for their contributions without which there would be no wonderful free dictionary and no free dataset either.

\bibliographystyle{elex}
\bibliography{elex}

\nocite{*}

\elexcopy

\end{document}